\documentclass{article} 
\usepackage{iclr2021_conference,times}

\usepackage{amsmath,amsfonts,bm}

\def\eqref#1{equation~\ref{#1}}

\def\1{\bm{1}}

\DeclareMathAlphabet{\mathsfit}{\encodingdefault}{\sfdefault}{m}{sl}
\SetMathAlphabet{\mathsfit}{bold}{\encodingdefault}{\sfdefault}{bx}{n}

\usepackage{hyperref}
\usepackage{url}
\usepackage{caption,subcaption}
\usepackage{graphicx}
\usepackage{booktabs}
\usepackage{multicol,multirow}
\usepackage{wrapfig}
\usepackage{enumitem}

\title{What's new? Summarizing Contributions in Scientific Literature}

\author{Hiroaki Hayashi~\thanks{Work done during internship at Salesforce Research} \\
Carnegie Mellon University \\
\texttt{hiroakih@cs.cmu.edu} \\
\And \And
Wojciech Kry\'sci\'nski,  Bryan McCann, Nazneen Rajani, Caiming Xiong\\
Salesforce Research \\
\texttt{\{kryscinski, bmccann, nazneen.rajani, cxiong\}@salesforce.com} \\
}

\iclrfinalcopy 
\begin{document}
\maketitle
\begin{abstract}
    With thousands of academic articles shared on a daily basis, it has become increasingly difficult to keep up with the latest scientific findings.
    To overcome this problem, we introduce a new task of \textit{disentangled paper summarization}, which seeks to generate separate summaries for the paper contributions and the context of the work, making it easier to identify the key findings shared in articles.
    For this purpose, we extend the S2ORC corpus of academic articles, which spans a diverse set of domains ranging from economics to psychology, by adding disentangled  ``contribution'' and ``context'' reference labels.
    Together with the dataset, we introduce and analyze three baseline approaches: 1) a unified model controlled by input code prefixes, 2) a model with separate generation heads specialized in generating the disentangled outputs, and 3) a training strategy that guides the model using additional supervision coming from inbound and outbound citations.
    We also propose a comprehensive automatic evaluation protocol which reports the \textit{relevance}, \textit{novelty}, and \textit{disentanglement} of generated outputs.
    Through a human study involving expert annotators, we show that in 79\%, of cases our new task is considered more helpful than traditional scientific paper summarization.

\end{abstract}

\section{Introduction}\label{sec:introduction}
With the growing popularity of open-access academic article repositories, such as arXiv or bioRxiv, disseminating new research findings has become nearly effortless.
Through such services, tens of thousands of scientific papers are shared by the research community every month\footnote{\url{https://arxiv.org/stats/monthly_submissions}}.
At the same time, the unreviewed nature of mentioned repositories and the sheer volume of new publications has made it nearly impossible to identify relevant work and keep up with the latest findings.

Scientific paper summarization, a subtask within automatic text summarization, aims to assist researchers in their work by automatically condensing articles into a short, human-readable form that contains only the most essential information.
In recent years, abstractive summarization, an approach where models are trained to generate fluent summaries by paraphrasing the source article, has seen impressive progress.
State-of-the-art methods leverage large, pre-trained models~\citep{raffel2019t5, lewis2020bart}, define task-specific pre-training strategies~\citep{zhang2019pegasus}, and scale to long input sequences~\citep{zhao2020seal, zaheer2020bigbird}.
Available large-scale benchmark datasets, such as arXiv and PubMed~\citep{cohan2018discourse}, were automatically collected from online archives and repurpose paper abstracts as reference summaries.
However, the current form of scientific paper summarization where models are trained to generate paper abstracts has two caveats: 1) often, abstracts contain information which is not of primary importance, 2) the vast majority of scientific articles come with human-written abstracts, making the generated summaries superfluous.

To address these shortcomings, we introduce the task of disentangled paper summarization.
The new task's goal is to generate two summaries simultaneously, one strictly focused on the summarized article's novelties and contributions, the other introducing the context of the work and previous efforts. 
In this form, the generated summaries can target the needs of diverse audiences: senior researchers and field-experts who can benefit from reading the summarized contributions, and newcomers who can quickly get up to speed with the intricacies of the addressed problems by reading the context summary and get a perspective of the latest findings from the contribution summary.

For this task, we introduce a new large-scale dataset by extending the S2ORC~\citep{lo2020s2orc} corpus of scientific papers, which spans multiple scientific domains and offers rich citation-related metadata.
We organize and process the data, and extend it with automatically generated contribution and context reference summaries, to enable supervised model training. 
We also introduce three abstractive baseline approaches:
1) a unified, controllable model manipulated with descriptive control codes~\citep{fan2018controllable,keskar2019CTRL},
2) a one-to-many sequence model with a branched decoder for multi-head generation~\citep{luong2016mtl, bansal2018mts}, 
and 3) an information-theoretic training strategy leveraging supervision coming from the citation metadata~\citep{peyrard2019simple}. 
To benchmark our models, we design a comprehensive automatic evaluation protocol that measures performance across three axes: relevance, novelty, and disentanglement.
We thoroughly evaluate and analyze the baselines models and investigate the effects of the additional training objective on the model's behavior.
To motivate the usefulness of the newly introduced task, we conducted a human study involving human annotators in a hypothetical paper-reviewing setting.
The results find disentangled summaries more helpful in 79\% of cases in comparison to abstract-oriented outputs.
Code, model checkpoints, and data preparation scripts introduced in this work are available at~\url{https://github.com/salesforce/disentangled-sum}.

\section{Related Work}\label{sec:related-work}

Recent trends in abstractive text summarization show a shift of focus from designing task-specific architectures trained from scratch~\citep{see2017get, paulus2018deeprl} to leveraging large-scale Transformer-based models pre-trained on vast amounts of data~\citep{liu2019text, lewis2020bart}, often in multi-task settings~\citep{raffel2019t5}.
A similar shift can be seen in scientific paper summarization, where state-of-the-art approaches utilize custom pre-training strategies~\citep{zhang2019pegasus} and tackle problems of summarizing long documents~\citep{zhao2020seal, zaheer2020bigbird}.
Other methods, at a smaller scale, seek to utilize the rich metadata associated with scientific articles and combine them with graph-based methods~\citep{yasunaga19scisumm}.
In this work, we combine these two lines of work and propose models that benefit from pre-training procedures, but also take advantage of task-specific metadata.

%
Popular large-scale benchmark datasets in scientific paper summarization~\citep{cohan2018discourse} were automatically collected from open-access paper repositories and consider article abstracts as the reference summaries.
Other forms of supervision have also been investigated for the task, including author-written highlights~\citep{collins2017supervised}, human annotations and citations~\citep{yasunaga19scisumm}, and transcripts from conference presentations of the articles~\citep{lev2019talksumm}.
In contrast, we introduce a large-scale automatically collected dataset with more fine-grained references than abstracts, which also offers rich citation-related metadata.

Update summarization~\citep{dang2008overview} defines a setting in a collection of documents with partially overlapping information is summarized, some of which are considered prior knowledge.
The goal of the task is to focus the generated summaries on the novel information.
Work in this line of research mostly focuses on novelty detection in news articles~\citep{bysani2010prog, delort2012dual} and timeline summarization~\citep{martschat2018tls, chang2016tls} on news and social media domains.
Here, we propose a novel task that is analogous to update summarization in that it also requires contrasting the source article with the content of other related articles which are considered pre-existing knowledge.

%
%

\section{Task}\label{sec:task}
Given a source article $D$, the goal of disentangled paper summarization is to simultaneously summarize the \textit{contribution}~$y_{con}$ and \textit{context}~$y_{ctx}$ of the source article.
Here, contribution refers to the novelties introduced in the article $D$, such as new methods, theories, or resources, while context represents the background of the work $D$, such as a description of the problem or previous work on the topic.
The task inherently requires a relative comparison of the article with other related papers to effectively disentangle its novelties from pre-existing knowledge.
%
Therefore, we also consider two sets of citations: inbound citations $C_I$ and outbound citations $C_O$ as potential sources of useful information for contrasting the article $D$ with its broader field.
%
Inbound citations refer to the set of papers that cite $D$, \textit{i.e.} relevant future papers, while outbound citations are the set of papers that $D$ cites, \textit{i.e.} relevant previous papers.
%
With its unique set of goals, the task of disentangled paper summarization poses a novel set of challenges for automatic summarization systems to overcome:
1) identifying salient content of $D$ and related papers from $C_I$ and $C_O$,
2) comparing the content of $D$ with each document from the citations,
and 3) summarizing the article along the two axes: contributions and context.

\subsection{Dataset}%
\label{sec:datasets}

Current benchmark datasets used for the task of scientific paper summarization, such as arXiv and PubMed~\citep{cohan2015scientific}, are limited in size, the number of domains, and lack of citation metadata. 
Thus, we construct a new dataset based on the S2ORC~\citep{lo2020s2orc} corpus, which offers a large collection of scientific papers spanning multiple domains along with rich citation-related metadata, such as citation links between papers and annotated citation spans.
Specifically, we carefully curate the data available in the S2ORC corpus and extend it with new reference labels.
%
%
\paragraph{Data Curation}
Some papers in the S2ORC corpus\footnote{\scriptsize Release ID: 20190928.} do not contain a complete set of information required by our summarization task: paper text, abstract, and citation metadata.
We remove such instances and construct a paper summarization dataset in which each example a) has an abstract and body text, and b) has at least 5 or more inbound and outbound citations, $C_I$ and $C_O$ respectively.
%
%
In cases where a paper has more than 20 incoming or outgoing citations, we sort them in descending order by the number of their respective citation and keep the top 20 most relevant articles.

\paragraph{Citation Span Extraction}
Each article in the set of inbound and outbound citations can be represented by its full text, abstract, or the span of text associated with the citation.
In this study, we follow~\citet{qazvinian2008scientific} and~\citet{cohan2015scientific} in representing citations with the sentences in which the citation occurs.\footnote{\scriptsize If a publication is cited multiple times within a source article we concatenate all relevant sentences.}
Thus, an outbound citation is represented by a sentence from the source paper.
Usually, such sentences directly refer to the cited paper and place its content in relation to the source paper.
Analogously, an inbound citation is represented by sentences from the citing paper and relates its content with the source paper.

\begin{table}[tb]
\scriptsize
\centering
\caption{\small Token length statistics on the training split of our dataset compared to existing scientific paper summarization datasets. Contribution summaries tend to be shorter than context summaries.}
\label{tab:datastats}
\begin{tabular}{@{}lrrrrrr@{}}
\toprule
    \multirow{2}{*}{Dataset}  & \multirow{2}{*}{\#Examples} & \multicolumn{5}{c}{Avg. \#Tokens}  \\ \cmidrule{3-7}
        &   & Paper $D$ & Inbound $C_I$ & Outbound $C_O$ & Contribution $y_{con}$ & Context $y_{ctx}$   \\ \midrule
ArXiv (Train)          & 203037                       & 4938  & -       & -        & \multicolumn{2}{c}{220 (Total summary)} \\
PubMed (Train)         & 119924                       & 3016  & -       & -        & \multicolumn{2}{c}{203 (Total summary)} \\ \midrule
Ours - Train           & 805152                       & 6351  & 925     & 877      & 136           & 236     \\ 
\phantom{Ours -} Valid & 36129 &  6374 & 922 & 875 & 135 & 236  \\
\phantom{Ours -} Test  & 54242 &  6350 & 927 & 892 & 136 & 237  \\ \bottomrule
\end{tabular}
\end{table}


\paragraph{Reference Generation}
Our approach relies on the availability of reference summaries for both contributions and contexts.
However, such annotations are not provided or easily extractable from the S2ORC corpus, and collecting expert annotations is infeasible due to the associated costs.
%
Therefore, we apply a data-driven approach to automatically extract contribution and context reference summaries from the available paper abstracts.
First, we manually label 400 abstracts sampled from the training set.
Annotations are done on a sentence-level with binary labels indicating \textit{contribution-} and \textit{context-}related sentences\footnote{\scriptsize Sentences not labeled as contribution are considered context, we leave finer-grained labels for future work.}.
This procedure yields 3341 sentences with associated binary labels, which we refer to as golden standard references.
%
Next, we fine-tune an automatic sentence classifier using the golden standard data.
As our classifier we use SciBERT~\citep{beltagy2019scibert}, which after fine-tuning achieves 86.3\% accuracy in classifying \textit{contribution} and \textit{context} sentences on a held-out test set.
%
Finally, we apply the fine-tuned classifier to generate reference labels for all examples in our dataset, which we refer to as silver standard references.

The statistics of the resulting dataset are shown in Table~\ref{tab:datastats}.

\section{Models}\label{sec:model}

\begin{figure}[tb]
\centering
    \includegraphics[width=\linewidth]{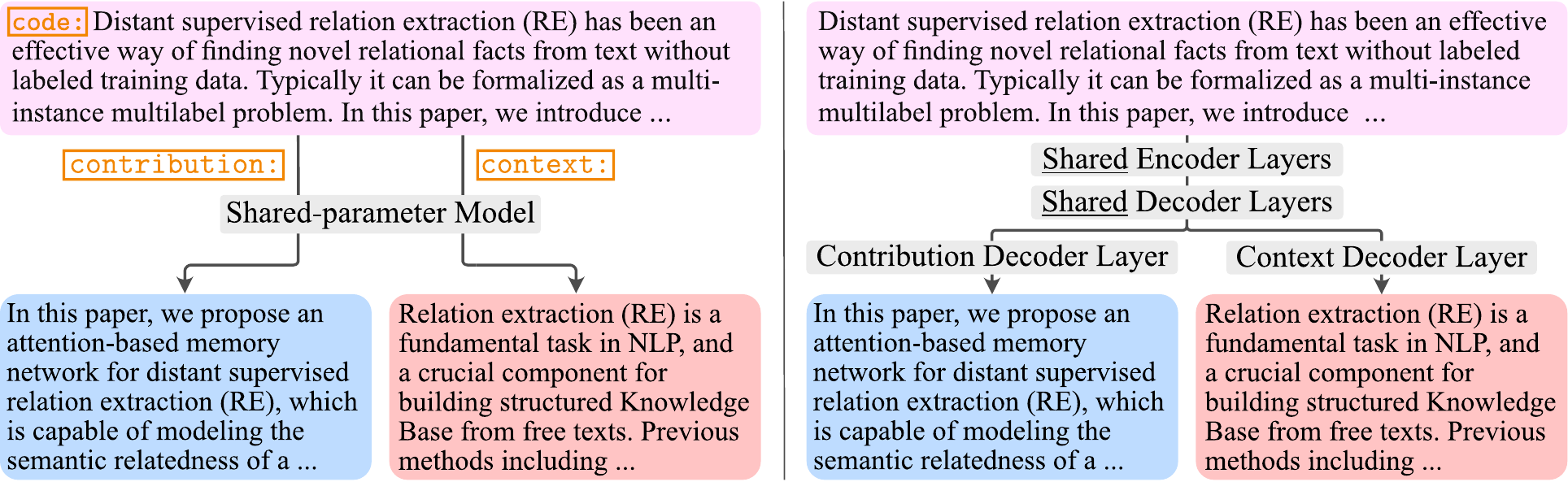}
    \caption{\small Model diagram. Left: \textsc{ControlCode} model, in which inputs are prefixed with a prompt symbol and passed to a shared model to control the output mode. Right: \textsc{MultiHead} model, which shares all of the model's parameters apart from the last decoder layer for different output modes, and chooses the final decoder layer accordingly to control the output mode.}
    \label{fig:model_diagram}
\end{figure}

Our goal is to build an abstractive summarization system which has the ability to generate contribution and context summaries based on the source article.
To achieve the necessary level of controllability, we propose two independent approaches building on encoder-decoder architectures:
%

\paragraph{\textsc{ControlCode} (CC)}
A common approach to controlling model-generated text is by conditioning the generation procedure on a control code associated with the desired output. 
Previous work on controllable generation~\citep{fan2018controllable,keskar2019CTRL} showed that prepending a special token or descriptive prompt to the model's input during training and inference is sufficient to achieve fine-grained control over the generated content.
Following this line of work, we modify our training instances by prepending textual control codes, \texttt{contribution:} or \texttt{context:}, to the summarized articles.
During training, all model parameters are updated for each data instance and the model is expected to learn to associate the provided prompt with the correct output mode.
The approach does not require changes in the architecture, making it straightforward to combine with existing large-scale, pre-trained models.
The architecture is shown on the left of Figure~\ref{fig:model_diagram}.
%
%

\paragraph{\textsc{MultiHead} (MH)}
An alternative way of controlling generation is by explicitly allocating layers within the model specifically for the desired control aspects.
Prior work investigating multi-task models~\citep{luong2016mtl, bansal2018mts} showed the benefits of combining shared and task-specific layers within a single, multi-task architecture.
Here, the encoder shares all parameters between the two generation modes, while the decoder shares all parameters, apart from the final layer, which splits into two generation branches. 
During training, each branch is individually updated with gradients from the associated mode.
%
The model shares the softmax layer weights between the output branches under the assumption that token-level vocabulary distributions are similar in the two generation modes due to the common domain. 
This approach is presented on the right of Figure~\ref{fig:model_diagram}.

\subsection{Informativeness-guided Training}
\label{sec:informativeness}

\cite{peyrard2019simple} proposed an information-theoretic perspective on text summarization which decomposes the criteria of a good summary into redundancy, relevance, and informativeness.
Among these criteria, \textit{informativeness} measures the user's degree of surprise after reading a summary given their background knowledge, and can be formally defined as:
\begin{align}
    \mathit{Inf}(D, K) = - \sum_i P_D(\omega_i) \log P_K(\omega_i),
\end{align}
where $\omega_i$ is a primitive semantic unit, $P_K$ is the probability over the unit under the user's knowledge, $P_D$ is the probability over the unit with respect to the source document, and $i$ is an index over all semantic units within a summary.


As defined by~\citet{peyrard2019simple}, informativeness is in direct correspondence to contribution summarization.
Paper contributions are novel contents introduced to the community, which causes surprisal given the general knowledge about the state of the field.
%
Therefore, in this work we explore utilizing this measure as an auxiliary objective that is optimized during training.
%
We define the semantic unit $\omega_i$ as the summary itself\footnote{\scriptsize For simplicity in modeling, we chose the entire summary.
However, this goes against the requirement set by \citet{peyrard2019simple} that $\omega_i$ is a \textit{primitive} semantic unit, because a paragraph's meaning can be decomposed into higher granular units.}
, which enables a simple interpretation of the corresponding probabilities.
We estimate $P_D$ as the likelihood of the summary given the paper content, $P_D(\omega_i) = p(y\,|\,D)$.
Since each paper is associated with a unique context and background knowledge, we treat the background knowledge as all relevant papers published before the source paper, \textit{i.e.}, outbound citations $C_O$.
Therefore, $P_K$ is estimated as the likelihood of the summary given the previous work, $P_K(\omega_i) = p(y \,|\, C_O)$.
We formulate the informativeness function as:
%
\begin{align}
    \label{eq:inf}
    \mathit{Inf}(D, K) = \begin{cases}
                - p(y_{con} \,|\, D) \log p(y_{con} \,|\,C_O) & \text{if generating contributions} \\
                - p(y_{ctx} \,|\, D) \log p(y_{ctx} \,|\,C_I) & \text{otherwise}
    \end{cases},
\end{align}
where the conditioning depends on the generation mode of the model, and aim to maximize it during the training procedure.
%

Combined with a cross entropy loss $L_{\mathit{CE}}$, we obtain the final objective which we aim to the minimize during training:
\begin{align}
\label{eq:loss}
    L = L_{\mathit{CE}} - \lambda \, \mathit{Inf}(D, K),
\end{align}
where $\lambda$ is a scaling hyperparameter determined through cross-validation.

\section{Experiments and Results}\label{sec:experiments-results}

In this section, we describe the experimental environment and report automatic evaluation results.
We consider four model variants:
\begin{itemize}
    \item \textbf{CC}, \textbf{CC+INF}: \textsc{ControlCode} model without and with the informativeness objective,
    \item \textbf{MH}, \textbf{MH+INF}: \textsc{MultiHead} model without and with the informativeness objective. 
\end{itemize}

\subsection{Evaluation}
\setlength\intextsep{0pt}
\begin{wrapfigure}[10]{r}[0pt]{0pt}
    \centering
    \includegraphics[width=0.45\linewidth]{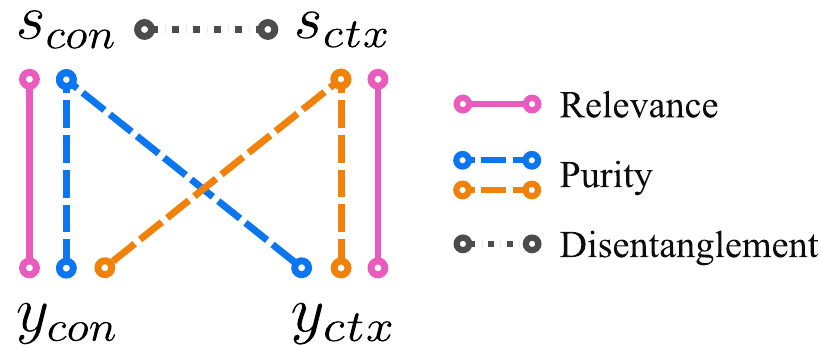}
    \caption{\small Diagram illustrating the evaluation protocol assessing summaries along 3 axes: relevance, purity, and disentanglement.}
    \label{fig:eval_diagram}
\end{wrapfigure}

\label{subsec:evaluation}
We perform automatic evaluation of the system outputs $(s_{con}, s_{ctx})$ against the silver standard references $(y_{con}, y_{ctx})$.
For this purpose, we have designed a comprehensive evaluation protocol, shown in Figure~\ref{fig:eval_diagram}, based on existing metrics that evaluates the performance of models across 3 dimensions:

\paragraph{Relevance} Generated summaries should closely correspond with the available reference summaries.
We measure the lexical overlap and semantic similarity between $(s_{con}, y_{con})$ and $(s_{ctx}, y_{ctx})$ using ROUGE (R-$i$)~\citep{lin2004rouge} and BERTScore (\citealt{zhang2020bertscore}; BS), respectively.

\paragraph{Purity} Generated contribution summary should closely correspond with its respective reference summary, but should not overlap with the context reference summary.
We measure the lexical overlap between $s_{con}$ and $(y_{con}$, $y_{ctx})$ using NouveauROUGE $_{con}$ (N$_{con}$-$i$)~\citep{conroy2011nouveau}.
The metric reports an aggregate score defined as a linear combination between the two components:
\begin{align*}
\text{NouveauROUGE}_{con}\text{-}i = \alpha^i_0 + \alpha^i_1 \text{ROUGE-}i(s_{con}, y_{con}) + \alpha^i_2 \text{ROUGE-}i(s_{con}, y_{ctx}),
\end{align*}
where weights $\alpha^i_j$ were set by the original authors to favor outputs with maximal and minimal overlap with related and unrelated references, accordingly.
Analogously, we calculate N$_{ctx}$-$i$ in reverse direction between $s_{ctx}$ and $(y_{ctx}$, $y_{con})$.
Purity P-$i$ is defined as the average novelty in both directions:
\begin{align*}
\text{Purity-}i = (\text{N}_{con}\text{-}i + \text{N}_{ctx}\text{-}i) / 2; & \text{\quad(P-}i\text{)}.
\end{align*}
\paragraph{Disentanglement} Generated contribution and context summaries should have minimal overlap.
We measure the degree of lexical overlap and semantic similarity between $(s_{con}, s_{ctx})$ using ROUGE and BERTScore, respectively.
To maintain consistency across metrics (higher is better) we report disentanglement scores as complements of the associated metrics:
\begin{align*}
\text{DisROUGE-}i = 100 - \text{ROUGE-}i; & \text{\quad(D-}i\text{)}, \\
\text{DisBERTScore}= 100 - \text{BERTScore}; & \text{\quad(DBS)}.
\end{align*}

\subsection{Implementation Details}

Our models build upon distilBART\footnote{\scriptsize We did not observe a substantial difference in performance between distilBART and BART.}~\citep{sanh2019distil, wolf2019hf}, a Transformer-based~\citep{vaswani2017attention}, pre-trained sequence-to-sequence architecture distilled from BART~\citep{lewis2020bart}.
Specifically, we used a model with 6 self-attention layers in both the Encoder and Decoder.
Weights were initialized from a model fine-tuned on a news summarization task.~\footnote{\scriptsize Model weights are available at \url{https://huggingface.co/sshleifer/student_cnn_6_6}.}
For the \textsc{MultiHead} model, the final layer of the decoder was duplicated and initialized with identical weights.
We fine-tuned on the training set for $80000$ gradient steps with a fixed learning rate of $3.0\times 10^{-5}$ and choose the best checkpoints in terms of ROUGE-1 scores on the validation set.
The loss scaling hyparameter $\lambda$ (Eq.~\ref{eq:loss}) was set to 0.05 and 0.01 for the \textsc{ControlCode} and \textsc{MultiHead} models, accordingly.
Input and output lengths were set to $1024$ and $200$, respectively.
At inference time, we decoded using beam search with beam size $5$.
The evaluation was performed using SummEval toolkit~\citep{fabbri2020summeval}.

\subsection{Results}
In Table~\ref{tab:results} we report results from the automatic evaluation protocol described in Subsection~\ref{subsec:evaluation}.

\paragraph{Relevance}
Across most models and metrics, relevance scores for context generation are higher than those for contribution summarization.
Manual inspection revealed that in some cases generated context summaries also include article contribution information, while this effect was not observed in the reverse situation.
Considering that silver standard annotations may contain noisy examples with incorrectly separated references, we suspect that higher ROUGE scores for context summaries may be caused by noisy predictions coinciding with noisy references.
Examples of such summaries are shown in the Appendix~\ref{app:samples}.
We also observe that informativeness-guided models (+INF) perform on par with their respective base versions, and the additional training objective does not affect the performance on the relevance metric.
This insight corroborates with~\citet{peyrard2019simple} who defines informativeness and relevance as orthogonal criteria.
\paragraph{Purity}
While the informativeness objective was designed to improve the novelty of generated summaries, results show an opposite effect, where informativeness-guided models slightly underperform their base counterparts.
The true reason for such behavior is unknown, however, it might be an indicator that the outbound citations $C_O$ are not a good approximation of reference context summaries $y_{ctx}$, or the relationship between the two is weak.
This effect is more evident in the Medical and Biology domains, which are the two most frequent domains in the dataset.

\paragraph{Disentanglement}
Results indicate that \textsc{ControlCode}-based models perform better than \textsc{MultiHead} approaches in terms of generating disentangled outputs.
This comes as a surprise given that the \textsc{CC} models share all parameters between the two generation modes, but might indicate that the two tasks contain complementary training signals.
We also noticed that, both informativeness-guided models performed better in terms of D-1.

Based on both purity and disentanglement evaluations, we suspect that the informativeness objective does guide the models to output more disentangled summaries (second term in Eq~\ref{eq:inf}), but the signal is not strong enough to focus on generating the appropriate content (first term in Eq~\ref{eq:inf}).
It is also clear that the \textsc{MultiHead} model benefits more from the additional training objective.
\begin{table}[tb]
\scriptsize
\centering
\caption{Automatic evaluation results on the test set. For all metrics, higher values indicate better results. Con and Ctx refer to contribution summary and context summary, respectively. Purity and Disentanglement are measaured on the pairs of contribution and context summaries.}
\begin{tabular}{@{}llrrrrrrrrrrrr@{}}
\toprule
\multirow{2}{*}{Model} &     & \multicolumn{4}{c}{Relevance}     & & \multicolumn{2}{c}{Purity} & & \multicolumn{4}{c}{Disentanglement}\\\cmidrule{3-6}\cmidrule{8-9}\cmidrule{11-14}
                       &     & R-1   & R-2   & R-L   & BS        & & P-1 & P-2                                 & & D-1 & D-2   & D-L   & DBS \\\midrule
\multirow{2}{*}{CC}    & Con & 39.16 & 15.96 & 24.65 & 63.22     & & \multirow{2}{*}{2.77} & \multirow{2}{*}{3.69} & & \multirow{2}{*}{52.95} & \multirow{2}{*}{72.18} & \multirow{2}{*}{69.12} & \multirow{2}{*}{33.62} \\
                       & Ctx & 41.84 & 17.24 & 24.55 & 63.78 \\\midrule
\multirow{2}{*}{CC+INF}& Con & 38.92 & 15.95 & 24.65 & 62.94     & & \multirow{2}{*}{2.75} & \multirow{2}{*}{3.68} & & \multirow{2}{*}{53.68} & \multirow{2}{*}{71.97} & \multirow{2}{*}{68.46} & \multirow{2}{*}{34.09} \\
                       & Ctx & 41.49 & 17.03 & 24.50 & 63.40 \\\midrule
\multirow{2}{*}{MH}    & Con & 39.20 & 15.98 & 24.72 & 63.04     & & \multirow{2}{*}{2.73} & \multirow{2}{*}{3.68} & & \multirow{2}{*}{50.89} & \multirow{2}{*}{69.51} & \multirow{2}{*}{65.97} & \multirow{2}{*}{32.51} \\
                       & Ctx & 41.67 & 17.23 & 24.65 & 63.77 \\\midrule
\multirow{2}{*}{MH+INF}& Con & 38.74 & 15.90 & 24.59 & 62.70     & & \multirow{2}{*}{2.68} & \multirow{2}{*}{3.60} & & \multirow{2}{*}{53.35} & \multirow{2}{*}{71.47} & \multirow{2}{*}{67.20} & \multirow{2}{*}{33.86} \\
                       & Ctx & 40.39 & 16.31 & 23.83 & 62.85 \\\bottomrule
\end{tabular}

\label{tab:results}
\end{table}

\section{Analysis}
\subsection{Qualitative Analysis}

To better understand the strengths and shortcomings of our models, we performed a qualitative study of model outputs.
Table~\ref{tab:samples} shows an example of generated summaries compared with the original abstract of the summarized article.
Our model successfully separates the two generation modes and outputs coherent and easy to follow summaries.
The contribution summary clearly lists the novelties of the work, while the context summary introduces the task at hand and explains its importance.
In comparison, the original abstract briefly touches on many aspects: the context, methods used, and contributions, but also offers details that are not of primary importance, such as the detailed about the simulation environment.

More generally, the described trends hold across summaries generated by our models.
The model outputs are fluent, abstractive, offer good separation between modes, and are on topic.
However, the factual correctness of summaries could not be assessed due to the highly specialized content and language of the summarized articles.
An artifact noticed in a few instances of the inspected outputs was leakage of contribution information into context summaries.
Other examples of generated summaries are included in the Appendix~\ref{app:samples}.
\begin{table}[tb]
\scriptsize
\centering
\caption{\small Generated samples compared with the original and generated abstracts of the associated paper.
The second rows shows the output decoded from DistilBART fine-tuned on our dataset, the third rows shows the outputs from \textsc{ControlCode} model.
Our model successfully generates disentangled content, thus making it easier to follow than the abstract.}
\begin{tabular}{@{}p{\textwidth}@{}}
\toprule
\textbf{Original Abstract:} Energy optimization in buildings by controlling the Heating Ventilation and Air Conditioning (HVAC) system is being researched extensively. In this paper, a model-free actor-critic Reinforcement Learning (RL) controller is designed using a variant of artificial recurrent neural networks called Long-Short-Term Memory (LSTM) networks. Optimization of thermal comfort alongside energy consumption is the goal in tuning this RL controller. The test platform, our office space, is designed using SketchUp. Using OpenStudio, the HVAC system is installed in the office. The control schemes (ideal thermal comfort, a traditional control and the RL control) are implemented in MATLAB. Using the Building Control Virtual Test Bed (BCVTB), the control of the thermostat schedule during each sample time is implemented for the office in EnergyPlus alongside local weather data. Results from training and validation indicate that the RL controller impoves thermal comfort by an average of 15\% and energy efficiency by an average of 2.5\% as compared to other strategies mentioned. \\ \midrule
\textbf{Generated Abstract:} Despite the advances in research on HVAC control algorithms, most field equipment is controlled using classical methods that include hysteresis/on/off and Proportional Integral and Derivative (PID) controllers. These classical methods do not perform optimally. The high thermal inertia of buildings induces large time delays in the building dynamics, which cannot be handled efficiently by the simple on/off controllers. However, due to the high non-linearity in building dynamics coupled with uncertainties such as weather, energy pricing, etc., these PID controllers require extensive retuning or auto-tuning capabilities, which increases the difficulty and complexity of the control problem. In this work, we introduce novel control algorithms from a branch of machine learning called reinforcement learning. From a controls perspective, reinforcement learning algorithms can be considered as direct adaptive optimal control. Like optimal control, reinforcement training algorithms minimize the cumulative sum of costs over a time horizon. Unlike traditional optimization algorithms can learn optimal control actions\\ \midrule
\textbf{Contribution:} In this work, we introduce novel control algorithms from a branch of machine learning called reinforcement learning. In our current approach, the impetus is thermostat control. Instead of traditional on/off heating and cooling control, reinforcement learning is utilized to set this schedule to obtain improved Predicted Mean Vote (PMV)-based thermal comfort at an optimal energy expenditure. Hence, a thermostats schedule is computed using an RL controller. The results show that the Q-learning algorithm can learn to adapt to time-varying and nonlinear system dynamics without explicit identification of the plant model in both systems and controls. \\
\textbf{Context:} The Heating, Ventilation and Air Conditioning (HVAC) systems can account for up to 50\% of total building energy demand. In the hopes of moving toward a greener, more energy-efficient future, a significant improvement in energy efficiency is needed to achieve this goal. Despite the advances in research on HVAC control algorithms, most field equipment is controlled using classical methods that include hysteresis/on/off and Proportional Integral and Derivative controllers. However, due to the high nonlinearity in building dynamics coupled with uncertainties such as weather, energy pricing, etc., these PID controllers require extensive retuning or auto-tuning capabilities, which increases the difficulty and complexity of the control problem. The high thermal inertia of buildings induces large time delays in the building dynamics, which cannot be handled efficiently by the simple on/off controllers. \\ \bottomrule
\end{tabular}
\label{tab:samples}
\end{table}

\subsection{Per-domain Performance}
Taking advantage of the rich metadata associated with the S2ORC corpus, we analyze the performance of models across the 10 most frequent scientific domains.
Table~\ref{tab:cc_dom} shows the results of contribution summarization using the \textsc{ControlCode}\footnote{\scriptsize The remaining models exhibit the same pattern.} model.

While ROUGE-1 scores oscillate around 40 points for most academic fields, the results indicate that summarizing documents from the Medical domain is particularly difficult, with models scoring about 7 points below average.

\begin{wraptable}[15]{r}{69mm}
\scriptsize
\centering
\caption{Relevance evaluation of contribution summaries for the top 10 domains generated using the \textsc{ControlCode} model. Performance on Medicine domain is paricularly low.}
\begin{tabular}{@{}lrrrr@{}}
\toprule
Metric            & R-1   & R-2   & R-L   & BS    \\\midrule
Biology           & 40.63 & 17.01 & 25.59 & 64.23 \\
Medicine          & 33.97 & 13.08 & 21.73 & 61.75 \\
Mathematics       & 40.13 & 15.56 & 24.42 & 61.58 \\
Computer science  & 43.54 & 16.41 & 25.86 & 63.43 \\
None              & 40.31 & 18.14 & 26.68 & 64.00 \\
Psychology        & 39.51 & 15.56 & 24.34 & 62.95 \\
Physics           & 40.09 & 15.85 & 24.89 & 62.10 \\
Chemistry         & 40.44 & 17.77 & 26.14 & 63.93 \\
Economics         & 39.56 & 14.25 & 23.41 & 60.91 \\
Materials science & 42.52 & 18.96 & 27.57 & 65.25 \\\bottomrule
\end{tabular}
\label{tab:cc_dom}
\end{wraptable}

Manual inspection of instances with low scores (R-1 $< 20$), exposed that contribution summaries in the Medical domain are highly quantitative (\textit{e.g.} ``Among these treated $\ldots$ retinopathy was noted in X\%'').
While other domains such as Biology also suffer from the same phenomenon, low-scoring quantitative summaries were 1.9 times more frequent in Medicine than in Biology.
An investigation into the domain distribution in our dataset (Appendix) revealed that Biology and Medicine are the two best represented fields in the corpus, with Biology having over twice as many examples.
We hypothesize that the poor performance of models stems from the fact that generating such quantitative summaries requires a deeper, domain-specific understanding of the source document and the available in-domain training data is insufficient to achieve that goal.

\subsection{Human Evaluation of Usefulness}
To assess the usefulness of the newly introduced task to the research community, we conducted a human study involving expert annotators.
The study aimed to compare disentangled papers summaries with traditional, abstract-based summaries in a hypothetical paper reviewing setting.
Judges were shown both types of summaries side by side and asked to pick one which would be more helpful for conducting the paper review.
Abstract-based summaries were generated by a model with a configuration identical to the models previously introduced in this work, trained to generate full abstracts using the same training corpus.
Annotators that participated in this study hold graduate degrees in technical fields and are active in the research community, however, they were not involved or familiar with this work prior to this experiment.

\begin{wraptable}[9]{r}{0.48\linewidth}
    \scriptsize
    \centering
        \caption{\small \label{tab:usefulness}Usefulness of disentangled summaries in percentage, \textit{e.g.}, Annotator 1 (A1) chose the disentangled summaries 82\% out of all the samples from S2ORC.}
    \begin{tabular}{@{}lrrrr@{}}
    \toprule
    Dataset  &  A1 & A2 & A3 & AVG. \\\midrule
    S2ORC & 82\% & 78\% & 70\% & 77\% \\
    CORD  & 88\% & 76\% & 78\% & 81\% \\\bottomrule
    \end{tabular}

\end{wraptable}

The study used 100 examples, out of which 50 were decoded on the test split of the adapted S2ORC dataset, while the other 50 were generated in a zero-shot fashion from articles in the CORD dataset~\citep{wang2020cord}, a recently introduced collection of papers related to COVID-19.
Results in Table~\ref{tab:usefulness} show the proportion of all examples where the annotators preferred the disentangled summaries over the generated abstracts.
The numbers indicate a strong preference from the judges for disentangled summaries, in the case of both S2ORC and CORD examples.
The values on CORD samples are slightly higher than those on S2ORC; we suspect this being due to the fact that the annotators were less familiar with the topics described in Covid-related publications and would require more help to review such articles. 
\section{Conclusions}\label{sec:conclusions}
In this paper, we propose \textit{disentangled paper summarization}, a new task in scientific paper summarizing where models simultaneously generate contribution and context summaries.
With the task in mind, we introduced a large-scale dataset with fine-grained reference summaries and rich metadata.
Along with the data, we introduced three abstractive baseline approaches to solving the new task and thoroughly assessed them using a comprehensive evaluation protocol design for the task at hand.
Through human studies involving expert annotators with motivated the usefulness of the task in comparison to the current scientific paper summarization setting.
Together with this paper, we release the code, trained model checkpoints, and data preprocessing scripts to support future work in this direction.
We hope this work will positively contribute to creating AI-based tools for assisting scientists in the research process.

\section{Acknowledgements}
The authors thank Wenhao Liu, Divyansh Agarwal, Sharvin Shah, and Tania Lopez-Cantu for assisting with annotations.

\bibliography{references}
\bibliographystyle{iclr2021_conference}

\clearpage
\appendix
\setlength\intextsep{12pt}
\section{Contribution Distribution in the Papers}
\label{app:contrib_dist}
\begin{figure}[tbhp]
    \centering
    \caption{Frequency of contribution mentions in different parts of papers. Around 22\% of sentences in the first bin are labeled as ``contribution''.}
    \includegraphics[width=0.45\linewidth]{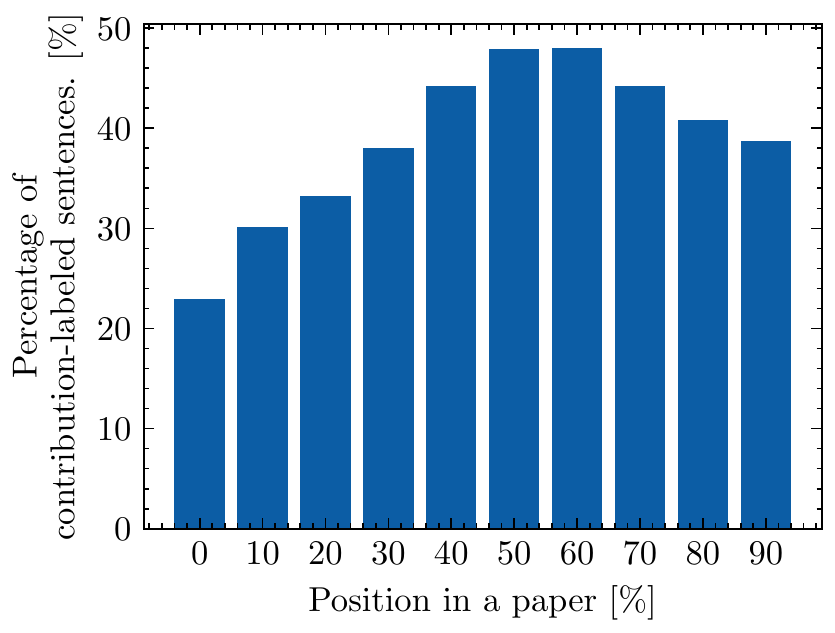}
    \label{fig:contrib_density}
\end{figure}

Different writing styles might locate and express contributions in different ways.
To understand the global tendency of contribution locations in a paper, we take each sentence from the paper texts themselves in the training set and annotated contributions using the learned sentence classifier.
We then group them into 10 bins according to the relative location of the sentences in the papers they belong to and constructed a distribution which summarizes the proportion of sentences labeled as contributions in each bin.
Fig~\ref{fig:contrib_density} shows the percentages of such sentences for each bin.
The graph shows that no bin positions in the papers tend to describe contributions more than 50\% of the time.
Surprisingly, the first 10\% of the papers have the lowest chance of describing the contributions, which is counter-intuitive to the general idea that papers tend to discuss the introduction and highlights of the paper at the beginning.

\section{Domain Distribution}

\begin{figure}[h]
    \centering
    \caption{Top-10 frequent domains in our dataset.}
    \includegraphics[width=0.45\linewidth]{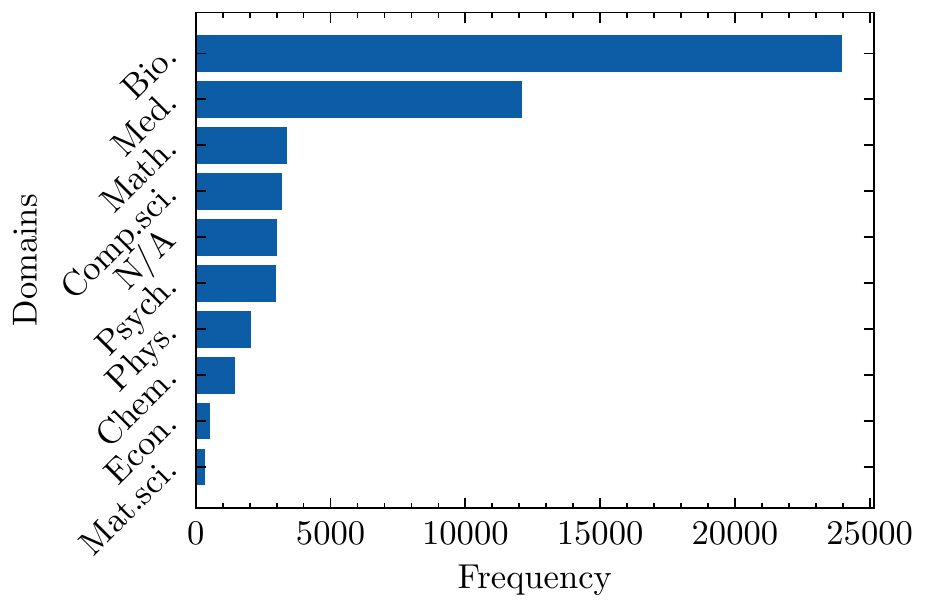}
    \label{fig:domainfreq}
\end{figure}

We show the Top-10 domain distribution of our dataset in Figure~\ref{fig:domainfreq}.
Biology and Medicine are the two most dominant domains.
5th most frequent ``domain'' is indicated as N/A, meaning that the domain information was not available by S2ORC.
Qualitatively, papers in the Biology domain tend to have a similarly formatted summary style to that of Medicine.

\section{Evaluation against Gold Annotations}
\label{app:gold_eval}
As discussed in Section~\ref{sec:datasets}, labels for contribution or context are populated automatically using a classifier, which is expected to contain mistakes.
Therefore, we created a gold standard evaluation set by manually annotating 100 samples in the test set and report the evaluation results in Table~\ref{tab:gold_disentanglement_small}.
A sharp drop in ROUGE scores for the context summaries is due to some examples receiving zero scores for generating context summaries when the manual annotation judged that there are not existent.
The overall trend of \textsc{ControlCode} model outpeforming \textsc{MultiHead} model is still observed in the evaluation.
More noticeably, we observe a reverse tendency when the two models are applied with the informativeness objective.
\textsc{MultiHead} model specifically enjoyed significant improvement in terms of novelty and disentanglement.
\begin{table}[tb]
\small
\centering
\caption{Automatic evaluation results on 100 samples from the test set with manual contribution annotations. For all metrics, higher values indicate better results.}
\begin{tabular}{@{}llrrrrrrrrrrrr@{}}
\toprule
\multirow{2}{*}{Model} &     & \multicolumn{4}{c}{Relevance}     & & \multicolumn{2}{c}{Purity} & & \multicolumn{4}{c}{Disentanglement}\\\cmidrule{3-6}\cmidrule{8-9}\cmidrule{11-14}
                       &     & R-1   & R-2   & R-L   & BS        & & P-1 & P-2                                & & D-1 & D-2   & D-L   & DBS \\\midrule
\multirow{2}{*}{CC}    & Con & 39.37 & 15.86 & 24.73 & 63.28     & & \multirow{2}{*}{2.30} & \multirow{2}{*}{3.22} & & \multirow{2}{*}{52.81}  & \multirow{2}{*}{71.52}  & \multirow{2}{*}{68.36} & \multirow{2}{*}{33.05} \\
                       & Ctx & 30.59 & 11.22 & 19.08 & 55.76 \\\midrule
\multirow{2}{*}{CC+INF}& Con & 38.38 & 15.21 & 23.47 & 62.59     & & \multirow{2}{*}{2.17} & \multirow{2}{*}{3.10} & & \multirow{2}{*}{52.49}  & \multirow{2}{*}{69.64}  & \multirow{2}{*}{66.60} & \multirow{2}{*}{32.76} \\
                       & Ctx & 30.14 & 11.10 & 19.00 & 55.55 \\\midrule
\multirow{2}{*}{MH}    & Con & 38.63 & 15.53 & 24.68 & 62.84     & & \multirow{2}{*}{2.21} & \multirow{2}{*}{3.13} & & \multirow{2}{*}{49.62}  & \multirow{2}{*}{67.45}  & \multirow{2}{*}{64.43} & \multirow{2}{*}{31.39} \\
                       & Ctx & 29.82 & 10.61 & 18.51 & 55.24 \\\midrule
\multirow{2}{*}{MH+INF}& Con & 39.43 & 15.75 & 24.77 & 63.11     & & \multirow{2}{*}{2.26} & \multirow{2}{*}{3.13} & & \multirow{2}{*}{51.56}  & \multirow{2}{*}{68.57}  & \multirow{2}{*}{64.97} & \multirow{2}{*}{32.35} \\
                       & Ctx & 29.14 & 10.25 & 18.48 & 54.92 \\\bottomrule
\end{tabular}
\label{tab:gold_disentanglement_small}
\end{table}

\clearpage
\setlength\intextsep{0pt}
\section{Human Evaluation of Disentanglement}
\label{app:humaneval_disentanglement}
\begin{wraptable}[11]{r}{35mm}
    \centering
    \caption{\label{tab:disentangle_bws}Disentanglement using Best-Worst scaling.}
    \begin{tabular}{@{}lr@{}}
    \toprule
    Model  & Rating \\\midrule
    CC     & \bf 0.027 \\
    CC+INF & 0.020 \\
    MH     & -0.073 \\
    MH+INF & \bf 0.027 \\\bottomrule
    \end{tabular}
\end{wraptable}

In addition to various automatic evaluation, we perform human evaluation on disentanglement to understand which models human annotators prefer.
We use Best-Worst scaling \citep{kiritchenko2017best} over the 4-tuples of summaries on the 50 random samples from the test set and have 3 annotators pick the best and the worst contribution and context summary pairs in terms of disentanglement.
The rating in Table~\ref{tab:disentangle_bws} shows the percentage a model is chosen as the best minus the percentage a model is chosen as the worst; rating ranges from -1 to 1.
A similar trend to automatic disentanglement evaluation is observed here as well in that introducing the informativeness objective significantly improves \textsc{MultiHead} model, while it conversely affects \textsc{ControlCode} model.

\section{Generated Full Samples from Different Models}
\label{app:samples}

We show additional samples generated from \textsc{ControlCode} model in Table~\ref{tab:other_samples2}, \ref{tab:other_samples3}, and \ref{tab:other_samples4}.
A common failure case of all our models is the lack of disentanglement between the two summaries.
While samples start generating on-topic, the model for context summary generation starts generating the contributions at one point.
This phenomenon can conspicuously be observed in Table~\ref{tab:other_samples2} and \ref{tab:other_samples4}.

\setlength\intextsep{12pt}
\begin{table}[htbp]
\small
\centering
\caption{Generated Sample. In this example, the red highlighted segments describe the contribution.}
\begin{tabular}{@{}p{\textwidth}@{}}
\toprule
\textbf{Original Abstract:} Domain adaption (DA) allows machine learning methods trained on data sampled from one distribution to be applied to data sampled from another. It is thus of great practical importance to the application of such methods. Despite the fact that tensor representations are widely used in Computer Vision to capture multi-linear relationships that affect the data, most existing DA methods are applicable to vectors only. This renders them incapable of reflecting and preserving important structure in many problems. We thus propose here a learning-based method to adapt the source and target tensor representations directly, without vectorization. In particular, a set of alignment matrices is introduced to align the tensor representations from both domains into the invariant tensor subspace. These alignment matrices and the tensor subspace are modeled as a joint optimization problem and can be learned adaptively from the data using the proposed alternative minimization scheme. Extensive experiments show that our approach is capable of preserving the discriminative power of the source domain, of resisting the effects of label noise, and works effectively for small sample sizes, and even one-shot DA. We show that our method outperforms the state-of-the-art on the task of cross-domain visual recognition in both efficacy and efficiency, and particularly that it outperforms all comparators when applied to DA of the convolutional activations of deep convolutional networks.  \\ \midrule
\textbf{Generated Abstract:} The difficulty of securing an appropriate and exhaustive set of training data, and the tendency for the domain of application to drift over time, often lead to variations between the distributions of the training (source) and test (target) data. In Machine Learning this problem is labeled domain mismatch. Failing to model such a distribution shift may cause significant performance degradation. Domain adaptation (DA) techniques capable of addressing this problem of distribution shift have thus received significant attention recently [24]. The assumption underpinning DA is that, although the domains differ, there is sufficient commonality to support adaptation. Many approaches have modeled this commonality by learning an invariant subspace, or set of subspaces. These methods are applicable to vector data only, however. Applying these methods to structured high-dimensional representations (e.g., convolutional activations), thus requires that the data be vectorized first. Although this solves the algebraic issue, it does not solve the underlying problem. Tensor\\ \midrule
\textbf{Contribution:} To address these issues, we propose a novel approach termed Tensor-Aligned Invariant Subspace Learning (TAISL) to learn an invariant tensor subspace that is able to adapt the tensor representations directly. By introducing a set of alignment matrices, the tensors from the source domain are aligned to an underlying tensor space shared by the target domain. Instead of executing a holistic adaptation (where all feature dimensions would be taken into account), our approach performs mode-wise partial adaptation where each mode is adapted separately to avoid the curse of dimensionality. We also propose an alternating minimization scheme which allows the problem to be effectively optimized by off-the-shelf solvers. Extensive experiments on cross-domain visual recognition demonstrate the following merits of our approach: i) it effectively reduces the domain discrepancy and preserves the discriminative power of the original representations; ii) it is applicable to small sample size adaptation, even when there is only one source\\
\textbf{Context:} Deep convolutional neural networks (CNNs) represent the state-of-the-art method for a substantial number of visual tasks. The activations of such CNNs, and the interactions between them, are naturally represented as tensors, meaning that DA should also be applied using this representation. However, after vectorization, many existing approaches become sensitive to the scarcity of source data (compared to the number of dimensions) and noise in the labels. \textcolor{red}{The proposed direct tensor method uses much lower dimensional entities, thus avoiding these estimation problems. To address these issues we propose to learn an invariant tensor subspace that is able to adapt the tensor representations directly. We show in Section 5 that the proposed method outperforms all comparators in DA of the Convolutional Activation of CNNs. Higher-order tensor modeling offers us an opportunity to investigate multiple interactions and couplings that capture the commonality and differences between domains. Following this idea, a novel approach} \\ \bottomrule
\end{tabular}
\label{tab:other_samples2}
\end{table}


\begin{table}[htbp]
\small
\centering
\caption{Generated Sample. Context summary has a slight erroneous decoding of contribution information.}
\begin{tabular}{@{}p{\textwidth}@{}}
\toprule
\textbf{Original Abstract:} Conventional wisdom holds that model-based planning is a powerful approach to sequential decision-making. It is often very challenging in practice, however, because while a model can be used to evaluate a plan, it does not prescribe how to construct a plan. Here we introduce the\"Imagination-based Planner\", the first model-based, sequential decision-making agent that can learn to construct, evaluate, and execute plans. Before any action, it can perform a variable number of imagination steps, which involve proposing an imagined action and evaluating it with its model-based imagination. All imagined actions and outcomes are aggregated, iteratively, into a\"plan context\"which conditions future real and imagined actions. The agent can even decide how to imagine: testing out alternative imagined actions, chaining sequences of actions together, or building a more complex\"imagination tree\"by navigating flexibly among the previously imagined states using a learned policy. And our agent can learn to plan economically, jointly optimizing for external rewards and computational costs associated with using its imagination. We show that our architecture can learn to solve a challenging continuous control problem, and also learn elaborate planning strategies in a discrete maze-solving task. Our work opens a new direction toward learning the components of a model-based planning system and how to use them.\\ \midrule
\textbf{Generated Abstract:} The Imagination-based Planner (IBP) is a model-based agent which learns from experience all aspects of the planning process: how to construct, evaluate, and execute a plan. The IBP learns when to act versus when to imagine, and if imagining, how to select states and actions for aggregating the proposed actions and evaluations into a useful plan. Through training, it effectively develops a planning algorithm tailored to the target problem. The learned algorithm allows it to flexibly explore, and exploit regularities in, the state and action spaces. In two experiments we evaluated a continuous IBP implementation on a challenging continuous control task, and a discrete IBP in a maze-solving problem. A fully learnable plan-based planning agent for continuous control. An agent that learns to construct a plan via model-inspired imagination was trained to perform \"conditional\" and \"adaptive computation,\" which results in a dynamic computational graph recently [7] trained neural networks trained to\\ \midrule
\textbf{Contribution:} In this paper we introduce the Imagination-based Planner (IBP), a model-based planning agent which learns from experience all aspects of the planning process: how to construct, evaluate, and execute a plan. The IBP learns when to act versus when to imagine, and if imagining, how to select states and actions to evaluate which will help minimize its external task loss and internal resource costs. Through training, it effectively develops a planning algorithm tailored to the target problem. In two experiments we evaluated a continuous IBP implementation on a challenging continuous control task, and a discrete IBP in a maze-solving problem.\\
\textbf{Context:} Model-based planning involves proposing sequences of actions, evaluating them under a model of the world, and refining these proposals to optimize expected rewards. Several key advantages of model-free methods are that models support generalization to states not previously experienced, help express the relationship between present actions and future rewards, and can resolve states which are aliased in value-based approximations. These advantages are especially pronounced in problems with complex and stochastic environmental dynamics, sparse reward, and restricted trial-and-error experience. Yet even with an accurate model, planning is often very challenging because while a model can be used to evaluate a plan, it does not prescribe how to construct a plan. Existing techniques for model based planning are most effective in small-scale problems, often require background knowledge of the domain, and use pre-defined solution strategies. \textcolor{red}{Here we introduce the Imagination-based Planner (IBP), a model-based agent which learns from experience all aspects of the planning}\\ \bottomrule
\end{tabular}
\label{tab:other_samples3}
\end{table}


\begin{table}[htbp]
\small
\centering
\caption{Generated Sample. More than half of the context summary is describing contribution.}
\begin{tabular}{@{}p{\textwidth}@{}}
\toprule
\textbf{Original Abstract:} We provide a generalization of discriminative lexicalized shift reduce parsing techniques for phrase structure grammar to a wide range of morphologically rich languages. The model is efficient and outperforms recent strong baselines on almost all languages considered. It takes advantage of a dependency based modelling of morphology and a shallow modelling of constituency boundaries.\\ \midrule
\textbf{Generated Abstract:} This paper shows that an efficient lexicalized phrase structure parser -modelling both dependencies and morphology -already significantly improves parsing accuracy. But we also show that an additional modelling of spans and constituency provides additional robustness that contributes to yield state of the art results on almost all languages considered, while remaining quite quite efficient. Moreover given the availability of existing multi-view treebanks (Bhatt et al., 2009; Seddah et al. 2014), our proposed solution only requires a lightweight infrastructure to achieve multilin-gual parsing without requiring costly languaged dependent modifications such as feature engineering. This method was used for the construction of the dataset for the SPMRL 2013 shared task. We first review the properties of multiview treebanks for a number of morphologically rich languages for which they are annotated both for constituents and dependencies that have the property to be token-wise aligned. These double annotations are typically obtained by converting a constituency or dependency annotation into the other annotation\\ \midrule
\textbf{Contribution:} In this paper, we propose a lexicalized phrase structure parsing model that can effectively take morphology into account. We show that an efficient lexicalization of both dependencies and morphology significantly improves multilingual state-of-the art results on almost all languages considered, while remaining quite efficient. Moreover, given the availability of existing multi-view treebanks, our proposed solution only requires a lightweight infrastructure to achieve multilin-gual parsing without requiring costly feature engineering such as feature engineering.\\
\textbf{Context:} Most state of the art multilingual parsers are weighted by discriminative models. Most state-of-the-art multilingual parsing methods rely on lexicalized phrase structure parsing techniques, which have recently been shown to improve performance in a variety of languages including free word order languages like English or Chinese. \textcolor{red}{In this paper we show that an efficient lexicallyized parser - modelling both dependencies and morphology - already significantly improves parsing accuracy. But it also shows that an additional modelling of spans and constituency provides additional robustness that contributes to yield state-ofthe-art results on almost all languages considered, while remaining quite efficient. Moreover, given the availability of existing multi-view treebanks (Bhatt et al., 2009; Seddah et al. 2013), our proposed solution only requires a lightweight infrastructure to achieve multilin-gual parsing without requiring costly feature engineering such as feature engineering.}\\ \bottomrule
\end{tabular}
\label{tab:other_samples4}
\end{table}


\end{document}